\documentclass[10pt,twocolumn,letterpaper]{article}

\usepackage{iccv}
\usepackage{times}
\usepackage{epsfig}
\usepackage{graphicx}
\usepackage{amsmath}
\usepackage{amssymb}

\usepackage{commath}
\usepackage{relsize}
\usepackage{booktabs}
\usepackage{pifont}
\usepackage{multirow}
\usepackage{color, colortbl}
\usepackage{pgfplots}
\usepackage{makecell}
\usepackage[titletoc]{appendix}

\pgfplotsset{compat=newest}
\definecolor{ForestGreen}{rgb}{0.13, 0.55, 0.13}
\definecolor{Maroon}{rgb}{0.69, 0.19, 0.0}
\newcommand{\cmark}{\ding{51}}%
\newcommand{\xmark}{\ding{55}}%
\newcommand{\multicmark}{\textcolor{Maroon}{\multirow{2}{*}{\cmark}}}
\newcommand{\multixmark}{\textcolor{ForestGreen}{\multirow{2}{*}{\xmark}}}
\newcommand{\singlecmark}{\textcolor{Maroon}{\cmark}}

\newcommand{\Gd}{\rowcolor{gray!45}}

\newcommand\Ga{}

\usepackage[breaklinks=true,bookmarks=false,colorlinks]{hyperref}

\iccvfinalcopy 

\newcommand{\BG}[1]{\textcolor{blue}{BG: #1}}

\newcommand{\eat}[1]{}
\newcommand*\rot{\rotatebox{90}}

\def\iccvPaperID{2690} 

\definecolor{cyan}{RGB}{35,164,205}
\definecolor{pinegreen}{RGB}{65,151,24}
\definecolor{gold}{RGB}{253,161,59}
\definecolor{yellow}{RGB}{252,231,36}
\definecolor{teal}{RGB}{54,90,140}
\definecolor{purple}{RGB}{70,9,92}
\definecolor{myred}{RGB}{169,52,31}
\definecolor{myblue}{RGB}{0,90,184}
\definecolor{mygreen}{RGB}{33, 120, 50}

\ificcvfinal\fi

\begin{document}

\title{Domain Randomization and Pyramid Consistency: \\Simulation-to-Real Generalization without Accessing Target Domain Data}

\author{Xiangyu Yue\textsuperscript{1}, Yang Zhang\textsuperscript{2}, Sicheng Zhao\textsuperscript{1}, Alberto Sangiovanni-Vincentelli\textsuperscript{1}, \\Kurt Keutzer\textsuperscript{1}, Boqing Gong\textsuperscript{3}\\
\textsuperscript{1} University of California, Berkeley,
\textsuperscript{2} University of Central Florida,
\textsuperscript{3} Google \\
\tt\small \{xyyue,schzhao,alberto,keutzer\}@berkeley.edu, yangzhang4065@gmail.com, boqinggo@outlook.com
}

\maketitle

\ificcvfinal\thispagestyle{empty}\fi

\begin{abstract}
We propose to harness the potential of simulation for the semantic segmentation of real-world self-driving scenes in a domain generalization fashion. The segmentation network is trained without any data of target domains and tested on the unseen target domains. To this end, we propose a new approach of domain randomization and pyramid consistency to learn a model with high generalizability. First, we propose to randomize the synthetic images with the styles of real images in terms of visual appearances using auxiliary datasets, in order to effectively learn domain-invariant representations. Second, we further enforce pyramid consistency across different ``stylized'' images and within an image, in order to learn domain-invariant and scale-invariant features, respectively. Extensive experiments are conducted on the generalization from GTA and SYNTHIA to Cityscapes, BDDS and Mapillary; and our method achieves superior results over the state-of-the-art techniques. Remarkably, our generalization results are on par with or even better than those obtained by state-of-the-art simulation-to-real domain adaptation methods, which access the target domain data at training time.

\end{abstract}

\vspace{-5mm}

\section{Introduction}
Simulation has spurred growing interests for  training deep neural nets (DNNs) for computer vision tasks~\cite{xiazamirhe2018gibsonenv, carla,kolve2017ai2,yue2018lidar}. This is partially due to the community's recent exploration to embodied  vision~\cite{tobin2017domain,zhu2017target,bousmalis2018using}, in which the perception has to be embodied and purposive for an agent to actively perceive and/or navigate through a physical environment~\cite{das2018embodied, carla}. Moreover, training data generated by simulation is often low-cost and diverse, especially benefiting the tasks that otherwise need heavy human annotations (\textit{e.g.} semantic segmentation~\cite{fcn_in_the_wild, zhang2017curriculum, Hoffman_cycada2017}). Finally, in the case of autonomous driving, simulation can complement the insufficient coverage of real data by synthesizing rare events and scenes, such as construction sites, lane merges, and accidents. In summary, the promise of simulation is that one may conveniently acquire a large amount of labeled and diverse imagery from simulated environments. This scale is vital for training state-of-the-art deep convolutional neural networks (CNNs) with millions of parameters.

\begin{figure}[t]
  \centering
  \includegraphics[width=3.2in, trim={0cm 0cm 0cm 0cm}]{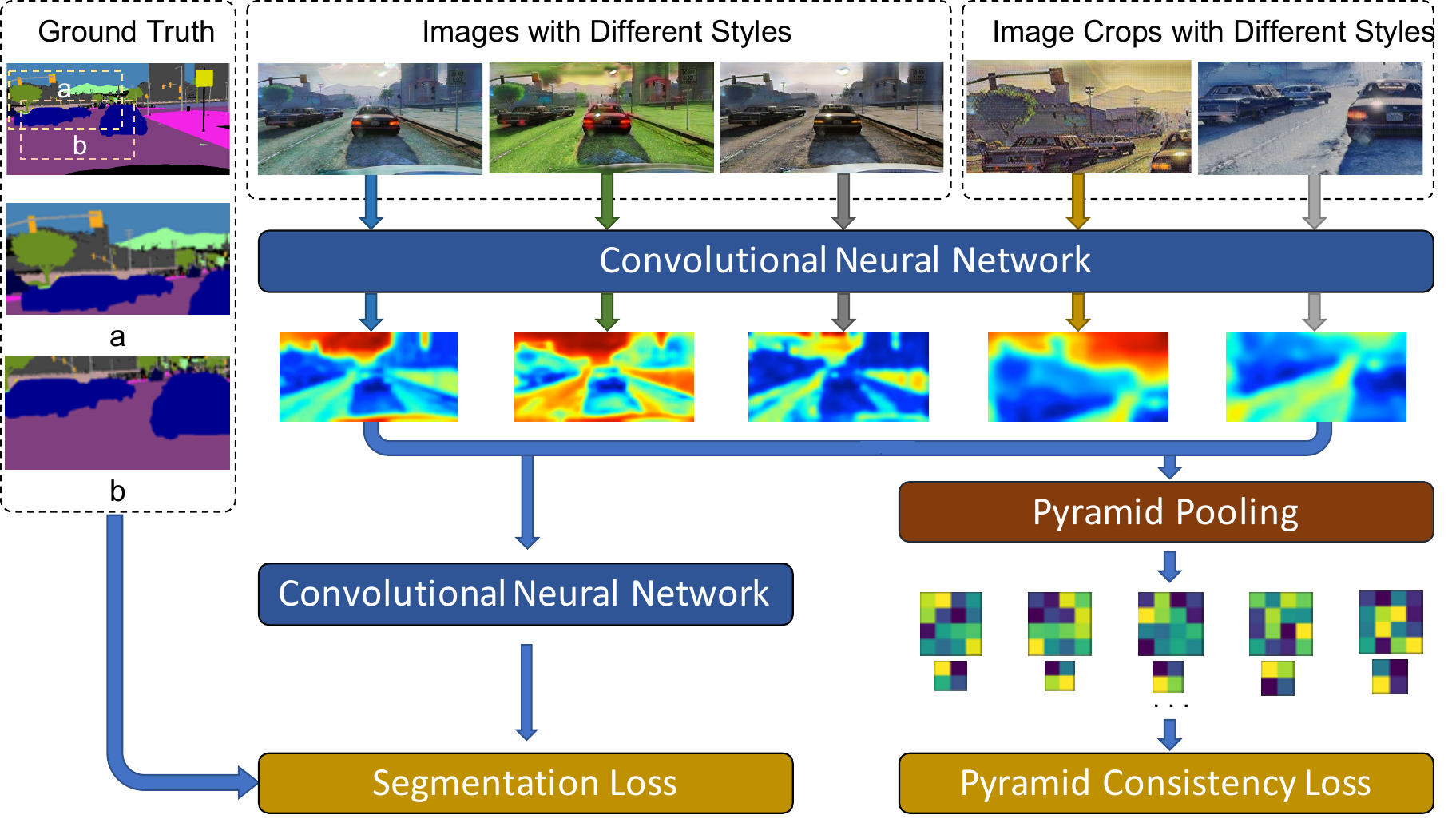}
  \caption{Domain randomization and pyramid consistency enforce the learned semantic segmentation network invariant to the change of domains. As a result, the semantic segmentation network can generalize to various domains, including those of real scenes.}
  \label{fig:teaser}
  \vspace{-10pt}
\end{figure}

However, when we learn a semantic segmentation neural network from a synthetic dataset, its visual difference from real-world scenes often discounts its performance on real images.  To mitigate the domain mismatch between simulation and the real world, existing work often resorts to \textbf{domain adaptation}~\cite{fcn_in_the_wild,Hoffman_cycada2017,zhang2017curriculum}, which aims to tailor the model for a particular target domain by jointly learning from the source synthetic data and the (often unlabeled) data of the target real domain. This setting is, unfortunately, very stringent. Take autonomous driving for instance. It is almost impossible for a car manufacturer to know in advance under which domain (which city, what weather, day or night) the vehicle would be used.

In this paper, we instead propose to harness the potential of simulation from a \textbf{domain generalization} manner~\cite{balaji2018metareg, li2018learning, ghifary2015domain, tobin2017domain}, without the need of accessing any target domain data in training and yet aiming to generalize well to multiple real-world target domains. We focus on the semantic segmentation of self-driving scenes, but the proposed method is readily applicable to similar tasks and scenarios. Our main idea is to randomize the labeled synthetic images to the styles of real images. We further enforce the semantic segmentation network to generate consistent predictions, in a pyramid form, over these domains. Our conjecture is that if the network is exposed to a sufficient number of domains in the training stage, it should  \emph{interpolate} well to new real-world target domains. In contrast, the domain adaptation work~\cite{fcn_in_the_wild, zhang2017curriculum,Hoffman_cycada2017} can be seen as \emph{extrapolating} from a single source domain to a single target domain. 

Our approach comprises two key steps: domain randomization and consistency-enforced training, as illustrated in Figure~\ref{fig:teaser}. Unlike~\cite{workshop_domain_randomization, sadeghi2016cad2rl}, we do not require any control of the simulators for randomizing the source domain imagery. Instead, we leverage the recently advanced image-to-image translation~\cite{CycleGAN2017} to transfer a source domain image to multiple styles, each dubbed an \textit{auxiliary domain}. This has at least three advantages over manipulating the simulator. First, it enables us to select auxiliary domains from the real world. After all, our goal is to achieve good performance on real data. Second, we have a more concrete anticipation about the look of the randomized images as we view the auxiliary domains. Finally, the randomized images are naturally grouped according to the auxiliary domains. The last point facilitates us to devise effective techniques in the second step to train the networks in a domain-invariant way. 

In the second step of our approach, we train a deep CNN for semantic segmentation with a pyramid consistency loss. If the network fits well to not only the synthetic source domain but also the auxiliary domains --- synthetic images with the styles of real images, it may become invariant to domain changes to a certain degree and thus generalize well to real-world target domain(s). To ensure consistent performance across different training domains, we explicitly regularize the network's internal activations so that they do not deviate from each other too much for the stylized versions of the same source domain image. We find that it is vital to apply the regularization over average-pooled pyramids rather than the raw feature maps, probably because the pooled pyramid gives the network certain flexibility.

To the best of our knowledge, this is the first work to explore domain randomization for the semantic segmentation problem. Experiments show that the proposed approach gives rise to robust domain-invariant CNNs trained using synthetic images. 
It significantly outperforms the straightforward source-only baseline and the newly designed network~\cite{pan2018twoatonce}, where the latter reduces the network's dependency on the training set by a hybrid of batch and instance normalizations. Our results are on par or even better than state-of-the-art domain adaptation results which are obtained by accessing the target data in training. 

\section{Related Work}
We now discuss some related work on semantic segmentation, domain adaptation, domain generalization, domain randomization, and data augmentation.

\eat{\BG{Remember to discuss related works accepted to CVPR 2019 --- some of them have become available on ArXiv. Especially, discuss~\cite{rui2018dlow} because it has also considered a generalization setting.}}

\textbf{Domain Adaptation for Semantic Segmentation.}
Until~\cite{fcn_in_the_wild,zhang2017curriculum} first studied the domain shift problem in semantic segmentation, most works in domain adaptation had focused on the task of image classification. After that, the problem subsequently became one of the tracks in the Visual Domain Adaptation Challenge (VisDA) 2017~\cite{peng2017visda} and started receiving increasing attention. Since then, adversarial training has been utilized in most of the following works~\cite{Hoffman_cycada2017, chen2018road, sankaranarayanan2018learning, fcan} for feature alignment. Most of these works were inspired by the unsupervised adversarial domain adaptation approach in~\cite{ganin2015unsupervised} which shares similar idea with generative adversarial networks. One of their most important objectives is to learn domain-invariant representations by trying to deceive the domain classifier. 
Zhang \textit{et al.}~\cite{zhang2017curriculum} perform segmentation adaptation by aligning label distributions both globally and across superpixels in an image. Recently, an unsupervised domain adaptation method has been proposed for semantic segmentation via class-balanced self-training~\cite{eccv_unsupervised}. Please refer to~\cite[Section 5]{zhang2019curriculum} for a brief survey of other related works.

\begin{figure*}[t]
  \centering
  \includegraphics[width=6.85in, trim={0.2cm 0cm 0cm 0cm}]{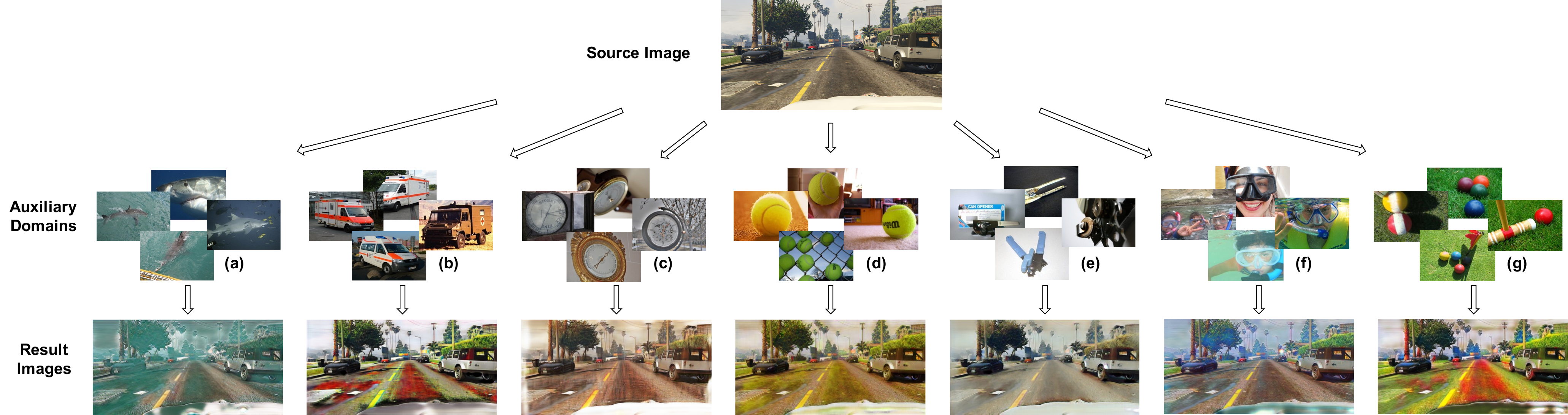}
  \caption{The domain randomization process. Top: an original synthetic image from the source domain; Mid: auxiliary image sets composed of ImageNet classes: (a) great white shark, (b) ambulance, (c) barometer, (d) tennis ball, (e) can opener, (f) snorkel, (g) tennis ball; Bottom: stylized images with same image content as the synthetic image and meanwhile corresponding styles of the ImageNet classes.}
  \label{fig:randomization}
  \vspace{-5pt}
\end{figure*} 

\textbf{Domain Generalization}
In contrast to Domain Adaptation, where the network is tested on a known target domain, and the images in the target domain, although without labels, are accessible during the training process, Domain Generalization is tested on unseen domains~\cite{muandet2013domain,gan2016learning}. Current domain generalization researches mostly focus on the image classification problem. Image data is hard to manually divide into discrete domains,~\cite{gong2013reshaping} devised a nonparametric formulation and optimization procedure to discover domains among both training and test data. \cite{li2018domain} imposed Maximum Mean Discrepancy measure to align the distributions among different domains and train the network with adversarial feature learning. \cite{li2017deeper} assigned a separate network duplication to each training domain during training and used the shared parameter for inference. \cite{li2018learning} improved generalization performance by using a meta-learning approach on the split training sets.

\textbf{Domain Randomization.}
Domain randomization (DR) is a complementary class of techniques for domain adaptation. Tobin \textit{et al.}~\cite{tobin2017domain} introduced the concept of Domain Randomization. Their approach randomly varies the texture and color of the foreground object, the background image, the number of lights in the scene, the
pose of the lights, the camera position, etc.  The goal is to close the reality gap by generating synthetic data with sufficient variation that the network views real-world data as just another variation. Randomization in the visual domain has been used to directly transfer vision-based policies from simulation to the real world without requiring real images during training~\cite{sadeghi2016cad2rl, tobin2017domain}. DR has also been utilized to do object detection and 6D pose estimation~\cite{nvidia_1,nvidia_2,6d_detection}. 
All the above DR methods require modifying objects inside the simulation environment. We instead propose a different DR method which is orthogonal to all the aforementioned methods. 

\textbf{Data Augmentation.}
Data augmentation is the process of supplementing a dataset with similar data created from the information in that dataset, which is ubiquitous in deep learning. When dealing with images, it often includes the application of rotation, translation, blurring, and other modifications~\cite{ciregan2012multicolumn, wan2013regularization, sato2015apac} to existing images that allow a network to better generalize~\cite{simard2003best}. In~\cite{lemley2017smart}, a network is proposed to automatically generate augmented data by merging two or more samples from the same class. A Bayesian approach is proposed in \cite{tran2017bayesian} to generate data based on the distribution learned from the training set. In~\cite{devries2017dataset},  simple transformations are used in the learned feature space to augment data. Counterexamples are considered to help data augmentation in ~\cite{dreossi2018counterexample}. Recently, AutoAugment has been proposed to learn augmentation policies from data~\cite{cubuk2018autoaugment}. The type of domain randomization we proposed in this paper can also be considered as a type of data augmentation.

\section{Approach}

The main idea of our approach is twofold, illustrated in Figure~\ref{fig:teaser}. The first part is \textbf{Domain Randomization with Stylization}: mapping the synthetic imagery to multiple auxiliary real domains (cf.\ Figure~\ref{fig:randomization}) in the training stage, such that, at the test stage, the target domain is not a surprise for the CNN model but merely another real domain. The second part is \textbf{Consistency-enforced Training}: enforcing pyramid consistency across domains and within an image to learn representations with better generalization ability.

\subsection{Domain Randomization with Stylization}
Keeping in mind that the target domain consists of real images, we randomly draw $K$ real-life categories from  ImageNet~\cite{deng2009imagenet}  for stylizing the synthetic images. Each category is called an auxiliary domain. We then use the image-to-image translation work~\cite{CycleGAN2017} to map the synthetic images to each of the auxiliary domains. As a result, the training set is augmented to $K+1$ times the original size.

\begin{figure*}[t]
  \centering
  \includegraphics[width=6.5in, trim={0cm 0cm 0cm 0cm}]{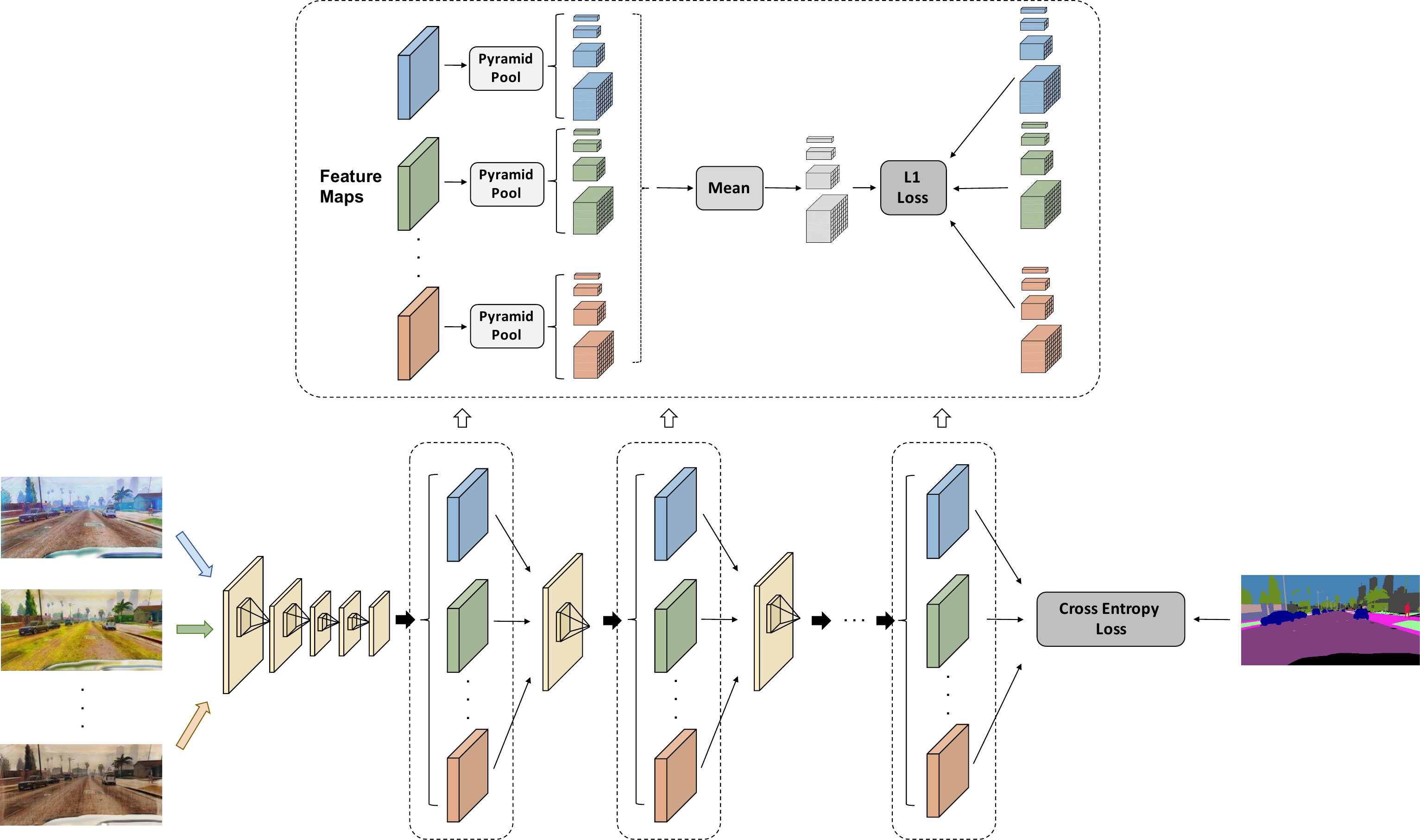}
  \vspace{2mm}
  \caption{Pyramid Consistency  across Domains. After feeding the images from different domains with the same content  into the neural network, we impose the pyramid consistency loss on the activation maps at each of the last few layers~(shown in blue, green and red).}
  \label{fig:consistency}
  
\end{figure*}

Figure~\ref{fig:randomization} illustrates this procedure and some qualitative results. We can see that each auxiliary domain stylizes the synthetic images by different real-world elements. Meanwhile, the semantic content of the original image is retained at most parts of the images. Some edge-preserving methods~\cite{li2017laplacian} on style transfer may give rise to better results, and are left for future work.

A straightforward method is to train a CNN segmentation model using the augmented training set. Denote by $\mathcal{D}^k, k=0,1,\cdots,K,$ the training domains, where $\mathcal{D}^0$ stands for the original source domain of synthetic images and $\mathcal{D}^k, k>0$ the auxiliary domains. A synthetic image $I_n^0\in \mathcal{D}^0$ has $K$ stylized copies $I_n^k\in \mathcal{D}^k$ in the auxiliary domains, and yet they all share the same semantic segmentation map $Y_n$ as the labels.  The objective function for training a  segmentation network $f(\cdot;\theta)$ is:
\begin{align}
\min_{\theta} \quad \mathcal{L}:= \frac{1}{Z}\sum_n \sum_{k=0}^{K} \,{L}\Big(Y_n,f(I_n^k;\theta)\Big), \label{eq:baseline}
\end{align}
where $\theta$ denotes the weights of the network,  ${L}(\cdot,\cdot)$ is the mean of pixel-wise cross-entropy losses, and $Z=(K+1)\abs{\mathcal{D}^0}$ is a normalization constant. 

Our experiments (cf.\ Section~\ref{sec:experiments_and_results}) show that the network trained using this augmented training set $\mathcal{D}^0\cup \mathcal{D}^1 \cdots \cup \mathcal{D}^K$ generalizes better to the unseen target domain than using the single source domain $\mathcal{D}^0$. Two factors may attribute to this result: 1) the training set is augmented in size and 2) the training set is augmented in style, especially in the styles closer to the real images. Despite being effective, this baseline method fails to track the multi-domain structure of the training set. We improve on it by the following.

\subsection{Consistency-enforced Training}
We aim to learn an image representation through the segmentation network that is domain-invariant for the semantic segmentation task. However, simply training the network with images from different domains (\textit{i.e.}, Eq.~(\ref{eq:baseline}), the baseline method) has some problems: a) images from different domains may drive the network toward distinct representations, making the training process not converge well; b) even if the network fits well to the training domains, it could capture the idiosyncrasies of each individual and yet fail at interpolating between them or to the new target domain.

In order to tackle these caveats, we  regularize the network by a consistency loss. The intuition is that if the network can generalize well, it should extract similar high-level semantic features and perform similar predictions for the images with the same content regardless of the styles.  The consistency loss is simply imposed as the following:

\begin{align}
\label{equ:consistent-domains}
\mathcal{R}:=\sum_{n,k}\sum_{l\in \mathcal{P}}\lambda_l\, L_{1}\Big(\overline{g_l(\mathcal{I}_n;\theta)},\, g_l(I_n^k;\theta)\Big),
\end{align}
where $l$ indexes an operator $g_l(\cdot;\theta)$ (\textit{e.g.}, average pooling) which maps a hidden layer's activations to a vector of smaller dimension, $\overline{g_l(\mathcal{I}_n;\theta)}$ denotes the target after each operation $g_l$ (cf.\ Sections~\ref{subsubsec:consistency_across_domains} and~\ref{subsubsec:consistency_within_image} for details), and $L_1$ is the $\ell_1$ distance. We argue that, by doing so, the network can be better guided to find a generic and domain-invariant representation for the semantic segmentation task.

The design of the operators $g_l(\cdot;\theta), l\in \mathcal{P}$ is key to the overall performance. The obvious identity mapping --- so the $\ell_1$ distance is directly calculated over the hidden activations --- does not work well in the experiments. One of the reasons is that it strictly requires the network to give about the same representations across different training domains, while some domains may be harder than the others to fit. 
\vspace{-2mm}
\subsubsection{Pyramid consistency across domains} 
\label{subsubsec:consistency_across_domains}
We find that the spatial pyramid pooling~\cite{he2014spatial, pspnet, pyramid_ori} serves as very effective operators $g_l(\cdot;\theta), l\in P$ in our context probably because it accommodates subtle differences of the network representations and meanwhile enables Eq.~(\ref{equ:consistent-domains}) to enforce the consistency at multiple scales. Pyramid pooling has been used in supervised visual understanding before, mostly as a part of the backbone networks. In this paper, instead, we use the pooled features to define regularization losses for training the network. The  pyramid consistency we consider is  over the images of different styles but with the same semantic content. 

Figure \ref{fig:consistency} illustrates our pyramid consistency scheme across the training domains. Consider a set of images $\mathcal{I}_n = \{I^k_n \mid k=0,1,\dots, K\}$ of $K+1$ different styles with the same annotation $Y_n$ and denote by  $M^{l,k}_n \in \mathbb{R}^{C_l \times H_l \times W_l}$ the feature map of input $I_n^k$ at layer $l$. Then, a spatial pyramid pooling operation is done on $M^{l,k}_n$. The spatial pyramid pooling operation is designed to fuse features under four different pyramid levels. First of all, a global average pooling is of the coarsest level that generates a single bin output. Each other pyramid levels separates the feature map into sub-regions evenly and performs average pooling inside each sub-region. In our design, we use $1\times 1$, $2\times2$, $4\times4$ and $8\times8$ as the pyramid pooling scales, namely the spatial size of the outputs of the pyramid pooling. After the pooling, we squeeze and concatenate the output tensors into a tensor $P^{l,k}_n \in \mathbb{R}^{C_l\times\left(1+2^2+4^2+8^2\right)}$, which is much lower-dimensional than the original feature map $M_n^{l,k}$. For a pair of images $I_n^{k}, I_n^{k'} \in \mathcal{I}_n$, the network is expected to have similar understanding and thus similar high-level features in a deep layer $l$. Note that simply constraining $M_n^{l,k}$ and $M_n^{l,k'}$ to be the same is too strong and could easily lead to degraded performance. To save computation, we avoided pair-wise terms and instead use the mean of $P_n^{l,k} (k=0,1,...,K)$ as the target value for the loss. Back to equation (\ref{equ:consistent-domains}), we have $g_l(I_n^k;\theta) = P_n^{l,k}$, the target is the mean across domains $\overline{g_l(\mathcal{I}_n;\theta)}=\frac{1}{K+1}\sum_k P_n^{l,k}$, and the set $\mathcal{P}=\{l\}$ is the layers down deep of the network. 

\eat{
get the $g_l$ and $\widehat{g_l}$ for the pyramid consistency across domains:
\begin{align}
&g_l(I_n^k;\theta) = P_n^{l,k}\\
&\widehat{g_l}(I_0,\cdots,I_K;\theta) = \frac{1}{K+1}\sum_{k=0}^K P_n^{l,k}
\end{align}
}

\vspace{-4mm}
\subsubsection{Pyramid consistency within an image}
\label{subsubsec:consistency_within_image}
The pyramid consistency loss across the training domains can guide the network to learn style-invariant features so that it can generalize well to the unseen target domains with different appearances. However, in many cases, style is not the only difference between domains. The view angles and parameters of cameras also lead to systematic domain mismatches in terms of the layout and scale of scenes. Take the \textit{focal length} parameter for instance. With different focal lengths, the same objects may be of different scales as the fields of view vary. 

In order to alleviate the issues above, we propose to further apply the pyramid consistency between random crops  and  full images. The idea is to artificially randomize the scale of the images and, therefore, guide the network to be robust to the domain gap incurred by the scene layouts and scales. Formally, following the notations in Section~\ref{subsubsec:consistency_across_domains}, each image $I_n^k$ of size $(H, W)$ is first randomly cropped at the same height-width ratio, with the top-left corner at $(h_n^k, w_n^k)$ and with the height $\overline{h_n^k}$. Then the crop is scaled back to the full image size, denoted as $C_n^k$, and finally fed to the network. Denote by $M^{l,k}_n$ and $MC^{l,k}_n \in \mathbb{R}^{C_l \times H_l \times W_l}$ the feature maps of the image $I_n^k$ and crop $C_n^k$ at layer $l$, respectively. Denote by $\overline{M^{l,k}_n}$ the part of $M^{l,k}_n$ corresponding to the crop. When there is no significant padding through the layers, then $\overline{M^{l,k}_n}$ is of shape ${C_l \times (\rho\cdot H_l) \times (\rho\cdot W_l)}$, where $\rho ={\overline{h_n^k}}/{h} $. 

We perform the spatial pyramid pooling on the cropped feature map $\overline{M^{l,k}_n}$ and the feature map $MC^{l,k}_n$ of the crop. The results are the same-size maps, $\overline{P_n^{l,k}}, PC_n^{l,k} \in \mathbb{R}^{C_l\times\left(1+2^2+4^2+8^2\right)}$. Back to Eq.~(\ref{equ:consistent-domains}), we have $g_l(I_n^k;\theta)=PC_n^{l,k}$ and the target vector is $\overline{g_l(\mathcal{I}_n;\theta)}=\overline{P_n^{l,k}}$.

\eat{
we can get the $L_{dist}$ for the pyramid consistency between random crops and full images:
\begin{align}
L_{dist}\Big(\widehat{g_l}(I_0,\cdots,I_K;\theta),\, g_l(I_n^k;\theta)\Big) = L_{dist}\Big(\Bar{P}_n^{l,k}, PC_n^{l,k}\Big)

\end{align}
}

\section{Experiments and Results}
\label{sec:experiments_and_results}
In this section, we describe the experimental setup and present  results on the semantic
segmentation generalization by learning from
synthetic data. Experimental analysis and comparison with
other methods are also provided.

\subsection{Experimental Settings}
It should be emphasized that our experiment setting is different from domain adaptation. Since domain adaptation aims to achieve good performance on a particular target domain, it requires unlabeled target domain data during training and also (sometimes) uses some labeled target domain images for validation. In contrast, our model is trained without any target domain data and is tested on unseen domains. 
\vspace{-8mm}
\paragraph{Datasets.}
In our experiments, we use GTA~\cite{richter2016playingfordata} and SYNTHIA~\cite{ros2016synthia} as the source domains and a small subset of ImageNet~\cite{deng2009imagenet} as well as datasets used in CycleGAN~\cite{CycleGAN2017} as the auxiliary domains for ``stylizing'' the source domain images. We consider three target domains of real-world images, whose official validation sets are used as our test sets:  Cityscapes~\cite{cordts2016cityscapes}, Berkeley Deep Drive Segmentation~(BDDS)~\cite{yu2018bdd100k}, and Mapillary~\cite{neuhold2017mapillary}.

GTA is a vehicle-egocentric image dataset collected in a computer game with pixel-wise semantic labels. It contains 24,966 images with the resolution $1914\times1052$. There are 19 classes which are compatible with other semantic segmentation datasets of outdoor scenes \textit{e.g.} Cityscapes. 

SYNTHIA is a large synthetic dataset with pixel-level semantic annotations. A subset, SYNTHIA-RAND-CITYSCAPES, is used in our experiments which contains 9,400 images with annotations compatible with Cityscapes.

Cityscapes contains vehicle-centric urban street images taken 
from some European cities. There are 5,000 images with pixel-wise annotations. The images have the resolution of $2048 \times 1024$ and are labeled into 19 classes.


BDDS contains thousands of real-world dashcam video frames with accurate pixel-wise annotations. It has a compatible label space with Cityscapes and the image resolution is $1280 \times 720$. The training, validation, and test sets contain 7,000, 1,000 and 2,000 images, respectively.

Mapillary contains street-level images collected from all around the world.
The annotations contain 66 object classes, but only the 19 classes that overlap with Cityscapes and GTA are used in our experiments. It has a training set with 18,000 images and a validation set with 2,000 images. 
\label{para:experimental_setup}

\begin{figure}
\centering
\begin{tikzpicture}[scale=0.96]
\begin{axis}[
    title={\textsc{Accuracy vs Auxiliary Domains}},
    xlabel={Number of Auxiliary Domains},
    ylabel={mIoU},
    ylabel style={yshift=-0.5em},
    xmin=0, xmax=16,
    ymin=24, ymax=36,
    xtick={0, 1, 3, 5, 7, 15},
    ytick={24,26,28,30, 32, 34, 36},
    legend pos=south east,
    legend style={nodes={scale=0.8}},
    ymajorgrids=true,
    grid style=dashed,
    width=240pt
]
 
    \addplot+[
    myblue,
    mark=triangle*,
    solid,
    mark options={solid},
    style=thick
    ]
    coordinates {
    (0,29.81)(1,32.15)(3,32.85)(5,32.75)(7,34.00)(15,34.64)
    };
    \addplot+[
    myred,
    mark=triangle*,
    solid,
    mark options={solid},
    style=thick
    ]
    coordinates {
    (0,24.59)(1,25.35)(3,27.72)(5,28.46)(7,29.04)(15,30.14)};
    \addplot+[
    mygreen,
    mark=triangle*,
    solid,
    mark options={solid},
    style=thick
    ]
    coordinates {
    (0,26.63)(1,28.34)(3,29.43)(5,30.21)(7,31.18)(15,31.64)
    };
    
    \addplot+[
    myblue,
    mark=square*,
    dashed,
    mark options={solid},
    style=thick
    ]
    coordinates {
    (0,29.81)(1,31.78)(3,32.28)(5,33.53)(7,33.47)(15,34.84)
    };
    \addplot+[
    myred,
    mark=square*,
    dashed,
    mark options={solid},
    style=thick
    ]
    coordinates {
    (0,24.59)(1,26.27)(3,27.14)(5,28.06)(7,29.235)(15,29.78)};
    \addplot+[
    mygreen,
    mark=square*,
    dashed,
    mark options={solid},
    style=thick
    ]
    coordinates {
    (0,26.63)(1,28.51)(3,29.57)(5,29.53)(7,31.08)(15,31.29)
    };
    \legend{Cityscapes-A, BDDS-A, Mapillary-A, Cityscapes-B, BDDS-B, Mapillary-B}
\end{axis}
\end{tikzpicture}
\caption{Accuracy of FCN8s-VGG16 with varying numbers of auxiliary domains. Two domain sets A and B are used. Models are trained on GTA and tested on Cityscapes, BDDS, and Mapillary. }
\label{fig:curve}
\vspace{-3mm}
\end{figure}
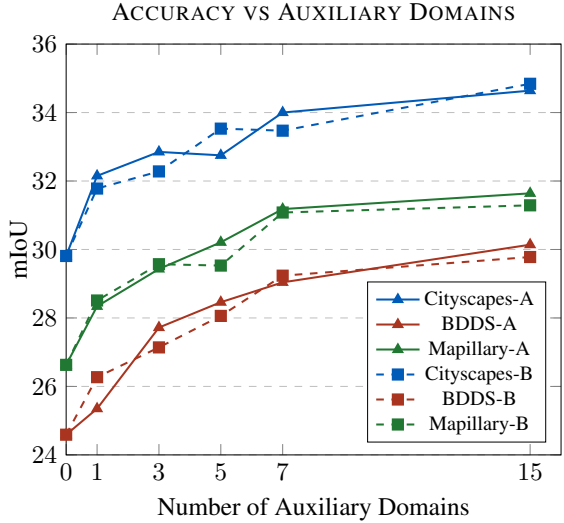

\vspace{-5mm}
\paragraph{Validation.}

To select a model for a particular real-world dataset $\mathcal{D}_R$ (\textit{e.g.} Cityscapes), we randomly pick up 500 images from the training set of another real-world dataset $\mathcal{D}_{R^{'}}$ (\textit{e.g.} BDDS) as the validation set. This cross-validation is to imitate the following real-life scenarios. When we train a neural network from a randomized source domain without knowing to which target domain it will be applied, we can probably collect a validation set which is as representative as possible of the potential target domains. Still take the car manufacturers for instance. A manufacturer may collect images of Los Angeles and NYC for the model selection while the cars will also be used in San Francisco and many  other cities. 
\vspace{-1mm}
\paragraph{Evaluation.} We evaluate the performance of a model on a test set using the standard PASCAL VOC intersection-over-union, \textit{i.e.} IoU.
The mean IoU~(mIoU) is the mean of all IoU values over all categories. To measure the generalizability of a model $M$, we propose a new metric,
\vspace{-1mm}
$$G_{perf}(M)=\mathbb{E}_{\mathcal{B}\in\mathcal{P}}\,mIoU(M, \mathcal{B}) \approx \frac{1}{L} \sum_{l} mIoU(M,\mathcal{B}_l)$$ 

where $\mathcal{B}$ is an unseen domain drawn from a distribution of all possible real-world domains $\mathcal{P}$, and $L$ is the number of unseen test domains, which is 3 in our experiment setting.

\paragraph{Implementation Details}
In our experiments, we choose to use FCN~\cite{long2015fully} as our semantic segmentation network. To make it easier to compare with most of other methods, we use VGG-16~\cite{vgg}, ResNet-50, and ResNet-101~\cite{he2016deep} as FCN backbones. The weights of the feature extraction layers in the networks are initialized from models trained on ImageNet~\cite{deng2009imagenet}. We add the pyramid consistency loss across domains on the last 5 layers, with $\lambda=0.2, 0.4, 0.6, 0.8, 1$, respectively. The pyramid consistency within an image is only added on the last layer. The network is implemented in PyTorch and trained with Adam optimizer~\cite{adam} using a batch size of 32 for the baseline models and 8 for our  models. Our machines are equipped with 8 NVIDIA Tesla P40 GPUs and 8 NVIDIA Tesla P100 GPUs.

\subsection{Evaluation of Domain Randomization}
\label{subsec:eval_domain_randomization}
In total, we use two sets of 15 auxiliary domains: A) 10 from ImageNet~\cite{deng2009imagenet} and 5 from CycleGAN~\cite{CycleGAN2017}, and B) 15 from ImageNet with each domain corresponding to one semantic class in Cityscapes. Please see supplementary materials for additional auxiliary domains, including color augmentation as an auxiliary domain.

To evaluate our domain randomization method, we conduct experiments generalizing from GTA to Cityscapes, BDDS, and Mapillary with FCN8s-VGG16. We augment the training set with images from different numbers of auxiliary domains in both setting A and B, and show the result in Figure~\ref{fig:curve}. As we can see from the plot, the accuracy increases with the number of auxiliary domains. The accuracy eventually saturates with the number of auxiliary domains. This is probably because 1) the 15 auxiliary domains are somehow sufficient to cover the appearance domain gap, and 2) as the number of images of the same content goes up, it is harder for the network to converge for the sake of the data scale and data variation. 

\begin{table}
\centering
\vspace{2mm}
\caption{Performance contribution of each design.}
\vspace{1mm}
\resizebox{\columnwidth}{!}{
\begin{tabular}{l|ccc|ccc}
\toprule
\multicolumn{1}{c|}{\multirow{2}{*}{\textbf{Method}}} & \multirow{2}{*}{\textbf{DR}} & \multirow{2}{*}{\textbf{PCD}} & \multirow{2}{*}{\textbf{PCI}} & \multicolumn{3}{c}{\textbf{mIoU}} \\ \cline{5-7} 
\multicolumn{1}{c|}{} & & & & Cityscapes & BDDS   & Mapillary \\ \midrule 
\Ga FCN & & & & 29.81 & 24.59   &26.63 \\ \hline
+DR & \cmark & & & 34.64 & 30.14   &31.64        \\ \hline
\Ga +PCD & \cmark & \cmark & & 35.47 & 31.21 &32.06  \\ \hline
+PCI & \cmark & & \cmark & 35.12 & 30.87 &32.12 \\ \hline
\Gd All & \cmark & \cmark & \cmark & \textbf{36.11} & \textbf{31.56}  & \textbf{32.25} \\ \bottomrule
\end{tabular}
}
\label{tab:ablation}
\end{table}

\begin{figure*}
  \centering
  \includegraphics[width=6.6in, trim={0cm 0cm 0cm 0cm}]{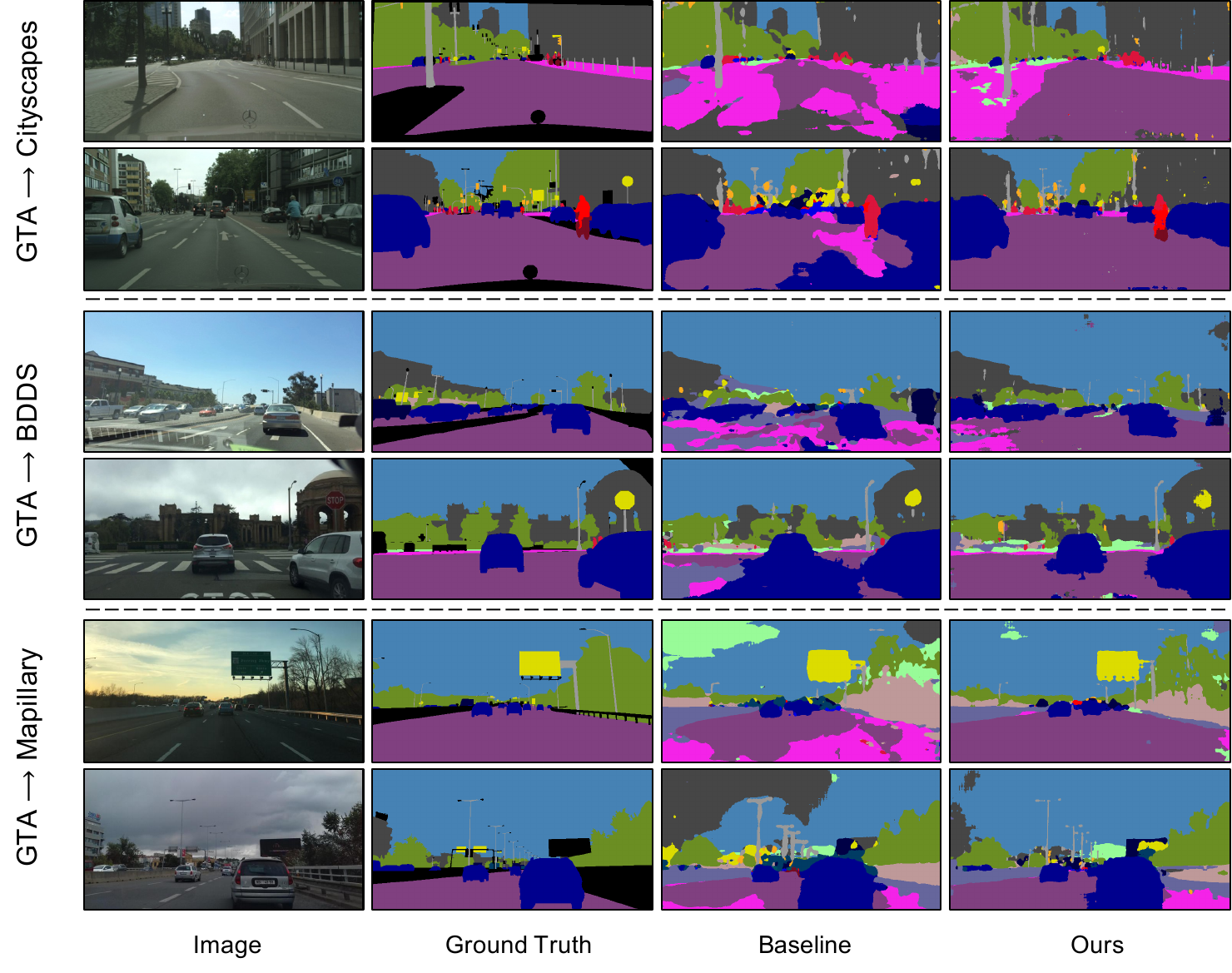}
  \caption{Qualitative semantic segmentation results of the generalization from GTA to Cityscapes, BDDS, and Mapillary.}
  \label{fig:visual}
\end{figure*}

\subsection{Ablation Study}
Next, we study how each design in our approach influences the overall performance. The experiments are still adapting from GTA to the 3 tests with FCN8s-VGG16.
Table~\ref{tab:ablation} details the mIoU improvement on Cityscapes, BDDS and Mapillary by considering one more factor each time: Domain Randomization~(DR), Pyramid Consistency across Domains~(PCD) and within an Image~(PCI). DR is a generic way to alleviate domain shift. In our case, it helps boost the performance from 29.81 to 34.64, from 24.59 to 30.14 and from 26.63 to 31.64, respectively for Cityscapes, BDDS and Mapillary. PCD and PCI further enhance the performance gains. By integrating all methods, our full approach finally reaches 36.11, 31.56 and 32.25 on Cityscapes, BDDS and Mapillary, respectively.
Figure~\ref{fig:visual} showcases some examples of the  semantic segmentation results on the 3 test sets.

\begin{table}

\centering
\caption{Domain generalization performance from (G)TA and (S)YNTHIA to (C)ityscapes, (B)DDS, and (M)apillary.}
\vspace{1mm}
\resizebox{\columnwidth}{!}{
\begin{tabular}{l|cc|cc|cc} \toprule
                                 & \multicolumn{2}{c}{VGG-16}  & \multicolumn{2}{c}{ResNet-50} & \multicolumn{2}{c}{ResNet-101} \\
                                 \cmidrule{2-7}
                                 & \multicolumn{1}{c}{NonAdapt} & \multicolumn{1}{c}{Ours} & \multicolumn{1}{c}{NonAdapt} & \multicolumn{1}{c}{Ours} & \multicolumn{1}{c}{NonAdapt} & \multicolumn{1}{c}{Ours} \\ \hline
 G$\rightarrow$ C   &29.81& \textbf{36.11}& 32.45& \textbf{37.42} & 33.56 & \textbf{42.53} \\
G$\rightarrow$ B   &24.59	&\textbf{31.56}	&26.73	&\textbf{32.14}	&27.76	&\textbf{38.72}  \\
 G$\rightarrow$ M  &26.63	&\textbf{32.25}	&25.66	&\textbf{34.12}	&28.33	&\textbf{38.05}\\
\Gd $G_{perf}$ &27.01	&\textbf{33.31}	&28.28	&\textbf{34.56}	&29.88	&\textbf{39.77} \\ \midrule
S $\rightarrow$ C &26.80	&\textbf{35.52}	&28.36	&\textbf{35.65}	&29.67	&\textbf{37.58} \\
S $\rightarrow$ B  &24.38	&\textbf{29.45}	&25.16	&\textbf{31.53}	&25.64	&\textbf{34.34} \\
S $\rightarrow$ M  &24.39	&\textbf{32.27}	&27.24	&\textbf{32.74}	&28.73	&\textbf{34.12} \\
\Gd $G_{perf}$  &25.34	&\textbf{32.41}	&26.92	&\textbf{33.31}	&28.01	&\textbf{35.35} \\  \bottomrule
\end{tabular}
}
\label{tab:domain_generalization}
\vspace{-4mm}
\end{table}

\begin{table}
\centering
\caption{Comparison with other domain generalization methods.}
\vspace{1mm}
\begin{tabular}{l|c|c|c} \toprule
Methods  & Base Net & mIoU & mIoU$\uparrow$ \\ \midrule
\Ga NonAdapt &    & 22.17 &  \\ 
\Ga IBN-Net~\cite{pan2018twoatonce} &\multirow{-2}{*}{ResNet-50} & 29.64 & \multirow{-2}{*}{7.47}\\
\Gd NonAdapt &     & 32.45   & \\ 
\Gd \textbf{Ours}     &   \multirow{-2}{*}{ResNet-50} & \textbf{37.42} & \multirow{-2}{*}{4.97} \\ 

\bottomrule
\end{tabular}
\vspace{-3mm}
\label{tab:compare_generalization}
\end{table}

\begin{table}[t]
\centering
\caption{Adaptation from GTA to Cityscapes with FCN-8s.}
\vspace{2mm}
\label{Tab: GTACityscapes}
\resizebox{\columnwidth}{!} {
\begin{tabular}{c|l||c|c|c|c}

\toprule
\hline
Network& Method & \multicolumn{1}{c|}{\begin{tabular}[c]{@{}c@{}}Train\\ w/\\ Tgt\end{tabular}} & \multicolumn{1}{c|}{\begin{tabular}[c]{@{}c@{}}Val\\ on\\ Tgt\end{tabular}} & mIoU & mIoU$\uparrow$ \\ \hline

\multicolumn{1}{l|}{\multirow{4}{*}{VGG-19}} & NonAdapt    & \multicmark  & \multicmark  & 22.3 & \multirow{2}{*}{6.6} \\
\multicolumn{1}{l|}{}                        & Curriculum~\cite{zhang2017curriculum}  & & & 28.9 & \\ \cline{2-6} 
\multicolumn{1}{l|}{}                        & NonAdapt    & \multicmark  & \multicmark & NA  & \multirow{2}{*}{NA} \\
\multicolumn{1}{l|}{}                        & CGAN~\cite{cgan}       &     &     & 44.5 &      \\ \hline
\multirow{30}{*}{VGG-16}                     & NonAdapt  & \multicmark   & \multicmark & 21.1 & \multirow{2}{*}{6.0} \\
                                             & FCN wld~\cite{fcn_in_the_wild}   &   &   & 27.1   &  \\ \cline{2-6} 
                                             & NonAdapt  & \multicmark & \multicmark & 17.9 & \multirow{2}{*}{17.5} \\
                                             & CYCADA~\cite{Hoffman_cycada2017}   &  & & 35.4  &  \\ \cline{2-6} 
                                             & NonAdapt  & \multicmark & \multicmark & 29.6   & \multirow{2}{*}{7.5}  \\
                                             & LSD~\cite{sankaranarayanan2018learning} &   &  & 37.1  &  \\ \cline{2-6} 
                                             & NonAdapt  & \multicmark  & \multicmark & 21.9 & \multirow{2}{*}{14.0} \\
                                             & ROAD~\cite{chen2018road}  &  &  & 35.9  & \\ \cline{2-6} 
                                             & NonAdapt & \multicmark & \multicmark & 24.9 & \multirow{2}{*}{3.9} \\
                                             & MCD~\cite{saito2017maximum} & &  & 28.8 & \\ \cline{2-6} 
                                             & NonAdapt & \multicmark & \multicmark & NA & \multirow{2}{*}{NA} \\
                                             & I2I~\cite{I2I}  &  &    & 31.8   &  \\ \cline{2-6} 
                                             & NonAdapt & \multicmark & \multicmark  & 24.3  & \multirow{2}{*}{11.8}  \\
                                             & CBST-SP~\cite{eccv_unsupervised} &  & & 36.1  &  \\ \cline{2-6} 
                                             & NonAdapt & \multicmark  & \multicmark  & 27.8 & \multirow{2}{*}{8.4}  \\
                                             & DCAN~\cite{wu2018dcan}  & &  & 36.2  &  \\ \cline{2-6} 
                                             & NonAdapt & \multicmark & \multicmark & 30.0 & \multirow{2}{*}{8.1} \\
                                             & PTP~\cite{zhu2018penalizing}  &  &  & 38.1 &  \\ \cline{2-6} 
                                             & NonAdapt & \multicmark & \multicmark & NA & \multirow{2}{*}{NA} \\
                                             & AdaptSegNet~\cite{tsai2018learning}  &  &  & 35.0 &  \\ \cline{2-6}
                                             & NonAdapt & \multicmark & \multicmark & NA & \multirow{2}{*}{NA} \\
                                             & BDL~\cite{li2019bidirectional}  &  &  & \textbf{41.3} &  \\ \cline{2-6} 
                                             & Non Adapt   &  \multicmark   & \multicmark     & 17.9 & \multirow{2}{*}{\textbf{18.7}} \\
                                             & CLAN~\cite{Yawei2019Taking}         &     &      & 36.6 & \\  \cline{2-6}
                                             & NonAdapt   & \multicmark & \multicmark & 18.8 & \multirow{2}{*}{13.8} \\
                                             & DAM~\cite{huang2018domain} &  &  & 32.6  & \\ \cline{2-6} 
                                             & NonAdapt & \multixmark & \multicmark & 30.0 & \multirow{2}{*}{8.6} \\
                                             & \textbf{Ours} & & & 38.6 & \\  \cline{2-6}
                                             & NonAdapt & \multixmark & \multixmark & 29.8 & \multirow{2}{*}{6.3} \\
                                             & \textbf{Ours} & & & 36.1 & \\ \hline
\end{tabular}
}

\end{table}

\begin{table}[t]
\centering
\caption{Adaptation from SYNTHIA to Cityscapes with FCN-8s.}
\vspace{2mm}
\label{Tab: SYNTHIA2Cityscapes}
\resizebox{\columnwidth}{!} {
\begin{tabular}{l|l||c|c|c|c}
\toprule
\hline
Network& Method & \multicolumn{1}{c|}{\begin{tabular}[c]{@{}c@{}}Train\\ w/\\ Tgt\end{tabular}} & \multicolumn{1}{c|}{\begin{tabular}[c]{@{}c@{}}Val\\ on\\ Tgt\end{tabular}} & mIoU & mIoU$\uparrow$ \\ \hline

\multicolumn{1}{c|}{\multirow{4}{*}{VGG-19}} & Non Adapt & \multicmark &  \multicmark     & 22.0 & \multirow{2}{*}{7.0} \\
\multicolumn{1}{c|}{}                        & Curriculum~\cite{zhang2017curriculum}   &   &       & 29.0                              &                               \\ \cline{2-6} 
\multicolumn{1}{c|}{}                        & Non Adapt    & \multicmark  & \multicmark & NA & \multirow{2}{*}{NA} \\
\multicolumn{1}{c|}{}                        & CGAN~\cite{cgan} &    &       & 41.2 & \\ \hline

\multirow{20}{*}{VGG-16}                      & Non Adapt & \multicmark & \multicmark & 17.4 & \multirow{2}{*}{2.8} \\
                                             & FCN Wld~\cite{fcn_in_the_wild} & & & 20.2 & \\ \cline{2-6} 
                                             & Non Adapt   & \multicmark    &  \multicmark     & 25.4 & \multirow{2}{*}{10.8} \\
                                             & ROAD~\cite{chen2018road} & & & 36.2 & \\ \cline{2-6} 
                                             & Non Adapt & \multicmark &  \multicmark & 26.8 & \multirow{2}{*}{9.3} \\
                                             & LSD~\cite{sankaranarayanan2018learning} & & & 36.1 & \\ \cline{2-6} 
                                             & Non Adapt & \multicmark & \multicmark & 22.6 & \multirow{2}{*}{\textbf{12.8}} \\
                                             & CBST~\cite{eccv_unsupervised} & & & 35.4 & \\ \cline{2-6} 
                                             & Non Adapt   & \multicmark & \multicmark & 27.8 & \multirow{2}{*}{8.4}   \\
                                             & DCAN~\cite{wu2018dcan} & & & 36.2 & \\ \cline{2-6} 
                                             & Non Adapt    & \multicmark  & \multicmark & NA & \multirow{2}{*}{NA} \\
                                             & DAM~\cite{huang2018domain} & & & 30.7 & \\ \cline{2-6} 
                                             & Non Adapt   & \multicmark   & \multicmark     & 24.9 & \multirow{2}{*}{9.3} \\
                                             & PTP~\cite{zhu2018penalizing} & & & 34.2 & \\  \cline{2-6} \cline{2-6}
                                             & Non Adapt   & \multicmark   & \multicmark     & NA & \multirow{2}{*}{NA} \\
                                             & BDL~\cite{li2019bidirectional}         &     &      & \textbf{39.0} & \\  \cline{2-6}
                                             & Non Adapt   &  \multixmark   & \multicmark & 27.3 & \multirow{2}{*}{9.1} \\
                                             & \textbf{Ours} &   & & 36.4 & \\ \cline{2-6}       
                                             & Non Adapt   &  \multixmark   & \multixmark & 26.8 & \multirow{2}{*}{8.7} \\
                                             & \textbf{Ours} & & & 35.5 & \\           
                                             
                                             \hline

\end{tabular}
}

\end{table}

\subsection{Generalization from GTA and SYNTHIA}
\label{subsec:generalization}
Then, we conduct extensive experiments to evaluate the generalization ability of our proposed methods. Specifically, we tested \textit{2 source domains}, GTA and SYNTHIA; \textit{3 models with different backbone networks}, VGG-16, ResNet-50 and ResNet-101; \textit{3 test sets}, Cityscapes, BDDS and Mapillary; and \textit{2 sets of auxiliary domains} (cf. Section~\ref{subsec:eval_domain_randomization}). The experiments with ResNet-50 are conducted with auxiliary domain set B, while the rest of the experiments are with set A. The validation set and test set in each experiment are from different domains, \textit{e.g.} using Cityscapes to select the model which will be evaluated on BDDS/Mapillary. 
The $G_{perf}$ value of each model is computed and the results are shown in Table~\ref{tab:domain_generalization}. We can see that the proposed techniques can greatly boost the generalizability by 5\%$\sim$12\% of different models regardless of dataset combinations. 

Then we compare our method with the only known existing state-of-the-art domain generalization method for semantic segmentation IBN-Net~\cite{pan2018twoatonce} under the generalization setting from GTA to Cityscapes. From the comparison shown in Table~\ref{tab:compare_generalization}, we can see that our domain generalization method has better final performance. IBN-Net improves domain generalization by fine-tuning the ResNet building blocks. Our method would be complementary with theirs.

\subsection{Adaptation from GTA and SYNTHIA}
\label{subsec:adaptation}
All experiments in the sections above are conducted in the domain generalization setting, where the validation set and the test set are from different domains. Now we conduct more experiments using the domain adaptation setting and compare our results with previous state-of-the-art works. Since most of the previous works conducted adaptation to Cityscapes with VGG backbone networks, we present the adaptation mIoU comparison on GTA $\rightarrow$ Cityscapes  and SYNTHIA $\rightarrow$ Cityscapes in Table~\ref{Tab: GTACityscapes} and Table~\ref{Tab: SYNTHIA2Cityscapes}, leaving class-wise comparison details in the supplementary material. We can see that our method is on par or better than the state-of-the-art methods in both settings. Further, we should notice that the domain generalization performance of our method~(last row) outperforms the adaptation performance of most other techniques.  In addition, since our method is target domain-agnostic, no data is needed from the target domain, resulting in more extensive applicability.

\section{Conclusion}
In this paper, we present a domain generalization approach for generalizing semantic segmentation networks from simulation to the real world without accessing any target domain data. We propose to randomize the synthetic images with auxiliary datasets and enforce pyramid consistency across domains and within an image. Finally, we experimentally validate our method on a variety of experimental settings, and show superior performance over state-of-the-art methods in both domain generalization and domain adaptation, which clearly demonstrates the effectiveness of our proposed method.

\textbf{Acknowledgement.} This work was partially supported by NSF grants, award 1645964, and by the Berkeley Deep Drive center. We thank Kostadin Ilov for providing system assistance.
{\small
\bibliographystyle{ieee_fullname}
\bibliography{mybib}
}

\clearpage
\appendix
\noindent\textbf{\Large Appendix}
\section{Detailed Comparison with Other Works.}

In Section~\ref{subsec:generalization} and Section~\ref{subsec:adaptation} of our main paper, we provide comparison of the overall performance (mean IoU) of the models, specifically comparison with other domain generalization works from GTA to Cityscapes, and with other domain adaptation works from GTA to Cityscapes as well as from SYNTHIA to Cityscapes. 
Here, we provide more detailed comparison of the class-wise accuracies in Table~\ref{tab:generalization}, Table~\ref{tab:adaptation_gta}, and Table~\ref{tab:adaptation_synthia}. From the detailed tables, we can see that our method provides better performance in many classes and outperforms the state-of-the-art methods in terms of \textit{mIoU} under both domain generalization and domain adaptation, which shows the efficacy and superiority of our method.

\section{Additional Experiments on auxiliary domains and color augmentation. }

Two more experiments are conducted with FCN8s-VGG16 in this section. First, we re-run our approach with 15 real-world styles from the BDD dataset, including different weather conditions, time of day~(TOD), \textit{etc.} Then, we replace the style transfer step with 15 color augmentations~\footnote{\url{https://github.com/aleju/imgaug}}, varying the hue, saturation,  grayscale, contrast, etc. These changes preserve the semantics of the objects.

Table~\ref{tab: perf} shows the new results (last two rows) along with those reported in the main paper.  ``Random'' stands for the styles randomly selected from ImageNet and Artworks, and ``Semantics'' are the styles of the Cityscapes classes~(\textit{e.g.}, Car, Road, etc.). The results are close to each other except that the color augmentation is a little worse than the others. The pyramid consistency  is effective for all the test cases.

\begin{table}[h]
\vspace{-3pt}
\centering
\caption{Adaptation from GTA with different style sets. We report results (mIoU\%) both without / with the pyramid  consistency.}
\vspace{1mm}
\label{tab: perf}
\resizebox{1\columnwidth}{!}{%
\begin{tabular}{l|c|c|c}
\hline
\multicolumn{1}{c|}{Style Set} & \multicolumn{1}{c|}{\begin{tabular}[c]{@{}c@{}}Semantics \\ Safe?\end{tabular}} &  \multicolumn{1}{c|}{Cityscapes} & \multicolumn{1}{c}{Mapillary} \\ \hline

Random & \xmark  & 34.64 / 36.11 & 31.64 / 32.25  \\ \hline 
Semantics & \xmark & 34.84 / 35.62 & 31.29 / 32.18 \\ \hline
Weather-TOD & \xmark & 34.51 / 35.89 & 31.24 / 32.18 \\  \hline
Color Change & \cmark& 33.56 / 34.52 & 30.27 / 32.06 \\ \hline 
\end{tabular}%
}
\vspace{-10pt}
\end{table}

\section{More Discussion.}
Table~\ref{tab: perf} shows that the color augmentation performs a little worse than the style transfers probably for two reasons. One is that it does not bring to the synthetic images any appearances of the real images by design. The other is that it randomizes the images only by color~(almost uniformly) and no texture. Learning an optimal non-uniform color shift policy is another future direction to explore.

Table~\ref{tab: perf} shows that different style sets, including the real styles~(\textit{i.e.} weather) suggested by R3, lead to similar results. Together with Figure 4 in the paper, we find that ``how many domains'' influences the results more than ``what  domains''.

\begin{table*}
\caption{Class-wise Performance comparison on Domain Generalization from GTA to Cityscapes with ResNet-50 base network.}
\vspace{1mm}

\resizebox{\textwidth}{!}{%
\begin{tabular}{c|l|cc|ccccccccccccccccccc|c}
\toprule[1.5pt]
\multirow{1}{*}[1.8em]{Network}&\multirow{1}{*}[1.8em]{Method} &\multirow{1}{*}[3em]{\makecell[bc]{Train \\ w/ \\ Tgt}} & \multirow{1}{*}[3em]{\makecell[bc]{Val \\ on \\ Tgt}} & \rot{road} & \rot{sidewalk} & \rot{building} & \rot{wall} & \rot{fence} & \rot{pole} & \rot{traffic light} & \rot{traffic sign} & \rot{vegettion} & \rot{terrain} & \rot{sky}   & \rot{person} & \rot{rider} & \rot{car}  & \rot{truck} & \rot{bus}  & \rot{train} & \rot{motorbike} & \rot{bicycle} & \rot{mIoU} \\ \hline

\multirow{4}{*}{ResNet-50} &NonAdapt~\cite{pan2018twoatonce} &\multixmark & \multixmark &-- &-- &-- &-- &-- &-- &-- &-- &-- &-- &-- &-- &-- &-- &-- &-- &-- &-- &-- & 22.17 \\
&IBN-Net~\cite{pan2018twoatonce}& & &-- &-- &-- &-- &-- &-- &-- &-- &-- &-- &-- &-- &-- &-- &-- &-- &-- &-- &-- & 29.64 \\ \cline{2-24}
&NonAdapt &\multixmark &\multixmark &84.5 &12.3 &75.4 &19.2 &9.1 &18.7 &19.2 &7.5 &81.6 &30.9 &73.8 &42.7 &8.9 &76.4 &17.2 &27.8 &1.8 &8.6 &1.2 &32.45\\
&Ours & & &90.1 &21.6 &79.4 &25.6 &18.2 &22.6 &26.4 &16.5 &82.9 &34.3 &77.1 &46.1 &13.5 &78.3 &24.4 &29.1 &3.6 &13.4 &7.8 &\textbf{37.42}\\ \bottomrule
\end{tabular}
}
\vspace{-2mm}
\label{tab:generalization}
\end{table*}

\begin{table*}
\caption{Class-wise Performance comparison from GTA to Cityscapes with VGG base network. All the best accuracies with respect to VGG-16 base network are in bold. }
\vspace{1mm}

\resizebox{\textwidth}{!}{%
\begin{tabular}{c|l|cc|ccccccccccccccccccc|c}
\toprule[1.5pt]
\multirow{1}{*}[1.8em]{Network}&\multirow{1}{*}[1.8em]{Method} &\multirow{1}{*}[3em]{\makecell[bc]{Train \\ w/ \\ Tgt}} & \multirow{1}{*}[3em]{\makecell[bc]{Val \\ on \\ Tgt}} & \rot{road} & \rot{sidewalk} & \rot{building} & \rot{wall} & \rot{fence} & \rot{pole} & \rot{traffic light} & \rot{traffic sign} & \rot{vegetation} & \rot{terrain} & \rot{sky}   & \rot{person} & \rot{rider} & \rot{car}  & \rot{truck} & \rot{bus}  & \rot{train} & \rot{motorbike} & \rot{bicycle} & \rot{mIoU} \\ \hline

\multirow{3}{*}{VGG19} &NonAdapt~\cite{zhang2017curriculum} & \multicmark & \multicmark & 18.1 & 6.8 & 64.1 & 7.3 & 8.7   & 21.0 & 14.9 & 16.8 & 45.9 & 2.4 & 64.4  & 41.6 & 17.5 & 55.3 & 8.4 & 5.0  & 6.9 & 4.3 & 13.8 & 22.3 \\ 
&Curriculum~\cite{zhang2017curriculum} & & & 74.9 & 22.0 & 71.7 & 6.0  & 11.9  & 8.4  & 16.3 & 11.1 & 75.7 & 13.3 & 66.5  & 38.0   & 9.3   & 55.2 & 18.8  & 18.9 & 0.0 & 16.8 & 16.6 & 28.9 \\ \cline{2-24}
&CGAN~\cite{cgan} & \singlecmark & \singlecmark &89.2 &49.0 &70.7 &13.5 &10.9 &38.5 &29.4 &33.7 &77.9 &37.6 &65.8 &75.1 &32.4 &77.8 &39.2 &45.2 &0.0 &25.5 &35.4 &44.5 \\ \hline

\multirow{26}{*}{VGG16} &NonAdapt~\cite{fcn_in_the_wild}& \multicmark & \multicmark & 31.9 & 18.9     & 47.7     & 7.4  & 3.1   & 16.0 & 10.4 & 1.0 & 76.5 & 13.0    & 58.9  & 36.0   & 1.0   & 67.1 & 9.5   & 3.7  & 0.0   & 0.0 & 0.0 & 21.1 \\ 
&FCNs Wld~\cite{fcn_in_the_wild} & &  & 70.4 & 32.4 & 62.1 & 14.9 & 5.4   & 10.9 & 14.2 & 2.7 & 79.2 & 21.3 & 64.6  & 44.1   & 4.2   & 70.4 & 8.0   & 7.3  & 0.0 & 3.5 & 0.0 & 27.1 \\ \cline{2-24}
&NonAdapt~\cite{sankaranarayanan2018learning} & \multicmark & \multicmark & 73.5 & 21.3 & 72.3     & 18.9 & 14.3  & 12.5 & 15.1 & 5.3 & 77.2 & 17.4 & 64.3  & 43.7   & 12.8  & 75.4 & 24.8  & 7.8  & 0.0   & 4.9 & 1.8 & 29.6 \\ 
&LSD~\cite{sankaranarayanan2018learning} &  &  & 88.0 & 30.5 & 78.6     & 25.2 & \textbf{23.5}  & 16.7 & 23.5 & 11.6 & 78.7 & 27.2 & 71.9  & 51.3   &19.5  & 80.4 & 19.8  & 18.3 & 0.9   & \textbf{20.8} & 18.4 & 37.1 \\ \cline{2-24}
&NonAdapt~\cite{chen2018road} & \multicmark  & \multicmark & 29.8 & 16.0 & 56.6 & 9.2  & 17.3  & 13.5 & 13.6 & 9.8 & 74.9 & 6.7 & 54.3  & 41.9 & 2.9 & 45.0 & 3.3   & 13.1 & 1.3   & 6.8 & 0.0 & 21.9 \\ 
&ROAD~\cite{chen2018road} & & & 85.4 & 31.2 & 78.6 &27.9 & 22.2  & 21.9 & 23.7 & 11.4 & 80.7 & 29.3 & 68.9  & 48.5   & 14.1  & 78.0 & 19.1  & 23.8 & \textbf{9.4}   & 8.3 & 0.0     & 35.9 \\ \cline{2-24}
&NonAdapt~\cite{Hoffman_cycada2017} & \multicmark & \multicmark & 26.0 & 14.9 & 65.1 & 5.5  & 12.9 & 8.9  & 6.0 & 2.5 & 70.0 & 2.9 & 47.0 & 24.5 & 0.0 & 40.0 & 12.1  & 1.5 & 0.0 & 0.0 & 0.0 & 17.9 \\ 
&CyCADA~\cite{Hoffman_cycada2017} &  & & 85.2 & 37.2 & 76.5 & 21.8 & 15.0  & 23.8 & 22.9 & \textbf{21.5} & 80.5 & 31.3 & 60.7  & 50.5   & 9.0   & 76.9 & 17.1  &28.2 & 4.5   & 9.8 & 0.0 & 35.4 \\ \cline{2-24}
&NonAdapt~\cite{saito2017maximum} & \multicmark & \multicmark & 25.9 & 10.9 & 50.5     & 3.3  & 12.2  & 25.4 &28.6 & 13 & 78.3 & 7.3 & 63.9  &52.1 & 7.9 & 66.3 & 5.2   & 7.8  & 0.9   & 13.7 & 0.7 & 24.9 \\ 
&MCD~\cite{saito2017maximum} & &  & 86.4 & 8.5 & 76.1     & 18.6 & 9.7   & 14.9 & 7.8 & 0.6 &82.8 & 32.7 & 71.4  & 25.2   & 1.1   & 76.3 & 16.1  & 17.1 & 1.4   & 0.2 & 0.0 & 28.8 \\ \cline{2-24}
&I2I~\cite{I2I} &\singlecmark   & \singlecmark & 85.3 & 38.0 & 71.3 & 18.6 & 16 & 18.7 & 12 & 4.5 & 72 & \textbf{43.4}    & 63.7  & 43.1   & 3.3   & 76.7 & 14.4  & 12.8 & 0.3 & 9.8 & 0.6 & 31.8 \\ \cline{2-24}

&NonAdapt~\cite{eccv_unsupervised} & \multicmark &\multicmark &64.0 &22.1 &68.6 &13.3 &8.7 &19.9 &15.5 &5.9 &74.9 &13.4 &37.0 &37.7 &10.3 &48.2 &6.1 &1.2 &1.8 &10.8 &2.9 &24.3 \\
&CBST-SP~\cite{eccv_unsupervised} & &  &\textbf{90.4} &\textbf{50.8} &72.0 &18.3 &9.5 &27.2 &28.6 &14.1 &82.4 &25.1 &70.8 &42.6 &14.5 &76.9 &5.9 &12.5 &1.2 &14.0 &\textbf{28.6} &36.1 \\ \cline{2-24}

&NonAdapt~\cite{wu2018dcan} &\multicmark &\multicmark &72.5 &25.1 &71.2 &6.6 &13.4 &12.3 &11.0 &4.7 &76.1 &16.4 &67.7 &43.1 &8.0 &70.4 &11.3 &4.8 &0.0 &13.9 &0.4 &27.8 \\ 
&DCAN~\cite{wu2018dcan} & & &82.3 &26.7 &77.4 &23.7 &20.5 &20.4 &\textbf{30.3} &15.9 &80.9 &25.4 &69.5 &52.6 &11.1 &79.6 &24.9 &21.2 &1.3 &17.0 &6.7 &36.2 \\ \cline{2-24}

&NonAdapt~\cite{Yawei2019Taking} &\multicmark &\multicmark &26.0 &14.9 &65.1 &5.5 &12.9 &8.9 &6.0 &2.5 &70.0 &2.9 &47.0 &24.5 &0.0 &40.0 &12.1 &1.5 &0.0 &0.0 &0.0 &17.9 \\ 
&CLAN~\cite{Yawei2019Taking} & & &88.0 &30.6 &79.2 &23.4 &20.5 &26.1 &23.0 &14.8 &81.6 &34.5 &72.0 &45.8 &7.9 &80.5 &26.6 &\textbf{29.9} &0.0 &10.7 &0.0 &36.6  \\ \cline{2-24}

&BDL~\cite{li2019bidirectional} &\singlecmark &\singlecmark &89.2 &40.9 &\textbf{81.2} &\textbf{29.1} &19.2 &14.2 &29.0 &19.6 &\textbf{83.7} &35.9 &\textbf{80.7} &\textbf{54.7} &\textbf{23.3} &\textbf{82.7} &25.8 &28.0 &2.3 &25.7 &19.9 &\textbf{41.3}  \\ \cline{2-24}

&NonAdapt~\cite{zhu2018penalizing} &\multicmark &\multicmark &-- &-- &-- &-- &-- &-- &-- &-- &-- &-- &-- &-- &-- &-- &-- &-- &-- &-- &-- &30.0\\
&PTP~\cite{zhu2018penalizing} & & &-- &-- &-- &-- &-- &-- &-- &-- &-- &-- &-- &-- &-- &-- &-- &-- &-- &-- &-- &38.1 \\ \cline{2-24}

&AdaptSeg~\cite{tsai2018learning} &\singlecmark &\singlecmark & 87.3 &29.8 &78.6 &21.1 &18.2 &22.5 &21.5 &11.0 &79.7 &29.6 &71.3 &46.8 &6.5 &80.1 &23.0 &26.9 &0.0 &10.6 &0.3 &35.0 \\ \cline{2-24}

&NonAdapt~\cite{huang2018domain} &\multicmark &\multicmark &-- &-- &-- &-- &-- &-- &-- &-- &-- &-- &-- &-- &-- &-- &-- &-- &-- &-- &-- &18.8\\
&DAM~\cite{huang2018domain} & & &-- &-- &-- &-- &-- &-- &-- &-- &-- &-- &-- &-- &-- &-- &-- &-- &-- &-- &-- &32.6 \\ \cline{2-24}

&NonAdapt & \multixmark& \multicmark & 68.4	&24.7 &68.9 &18.1 &15.2 &18.1 &16.7 &9.6 &78.4 &18.3 &65.7 &43.6 &12.3 &69.1 &18.7 &16.1 &0.4 &5.3 &3.2 &30.0 \\
&Ours &     &  &86.6 &38.4 &79.8 &26.4 &18.1 &\textbf{34.7} &21.3 &16.3 &81.2 &28.7 &76.5 &50.1 &16.6 &80.7 &\textbf{28.3} &21.4 &2.3 &14.3 &10.9 &38.6 \\ \cline{2-24}

&NonAdapt & \multixmark& \multixmark &66.4 &23.9 &69.1 &16.3 &15.8 &19.6 &15.8 &8.6 &77.7 &19.5 &66.1 &43.2 &12.1 &68.9 &17.3 &17.2 &0.3 &4.8 &2.9 &29.8\\ 
&Ours & &  &84.6 &31.5 &76.3 &25.4 &17.2 &28.2 &21.5 &13.7 &80.7 &26.8 &74.9 &47.5 &15.8 &77.1 &22.2 &22.7 &1.7 &8.9 &9.7 &36.1\\ \bottomrule
\end{tabular}
}
\label{tab:adaptation_gta}
\end{table*}

\begin{table*}
\caption{Class-wise Performance comparison from SYNTHIA to Cityscapes with VGG base network. All the best accuracies with respect to VGG-16 base network are in bold. }
\vspace{1mm}

\resizebox{\textwidth}{!}{%
\begin{tabular}{c|l|cc|cccccccccccccccc|c}
\toprule[1.5pt]
\multirow{1}{*}[1.8em]{Network}&\multirow{1}{*}[1.8em]{Method} &\multirow{1}{*}[3em]{\makecell[bc]{Train \\ w/ \\ Tgt}} & \multirow{1}{*}[3em]{\makecell[bc]{Val \\ on \\ Tgt}} & \rot{road} & \rot{sidewalk} & \rot{building} & \rot{wall} & \rot{fence} & \rot{pole} & \rot{traffic light} & \rot{traffic sign} & \rot{vegetation} & \rot{sky}   & \rot{person} & \rot{rider} & \rot{car} & \rot{bus} & \rot{motorbike} & \rot{bicycle} & \rot{mIoU} \\ \hline

\multirow{3}{*}{VGG19} &NonAdapt~\cite{zhang2017curriculum} & \multicmark & \multicmark &5.6 &11.2 &59.6 &0.8 &0.5 &21.5 &8.0 &5.3 &72.4 &75.6 &35.1 &9.0 &23.6 &4.5  &0.5 &18.0 &22.0 \\ 
&Curriculum~\cite{zhang2017curriculum} & & &65.2 &26.1 &74.9 &0.1 &0.5 &10.7 &3.7 &3.0 &76.1 &70.6 &47.1 &8.2 &43.2 &20.7 &0.7 &13.1 &29.0 \\ \cline{2-21}
&CGAN~\cite{cgan} & \singlecmark & \singlecmark &85.0 &25.8 &73.5 &3.4 &3.0 &31.5 &19.5 &21.3  &67.4 &69.4 &68.5 &25.0 &76.5 &41.6 &17.9 &29.5 &41.2 \\ \hline

\multirow{18}{*}{VGG16} &NonAdapt~\cite{fcn_in_the_wild}& \multicmark & \multicmark  &6.4 &17.7 &29.7 &1.2 &0.0 &15.1 &0.0 &7.2 &30.3 &66.8 &51.1 &1.5 &47.3 &3.9 &0.1 &0.0 &17.4 \\ 
&FCNs Wld~\cite{fcn_in_the_wild} & & &11.5 &19.6 &30.8 &4.4 &0.0 &20.3 &0.1 &11.7 &42.3 &68.7 &51.2 &3.8 &54.0 &3.2 &0.2 &0.6 &20.2 \\ \cline{2-21}
&NonAdapt~\cite{sankaranarayanan2018learning} & \multicmark & \multicmark  &30.1 &17.5 &70.2 &5.9 &0.1 &16.7 &9.1 &12.6 &74.5 &76.3 &43.9 &13.2 &35.7 &14.3 &3.7 &5.6 &26.8 \\ 
&LSD~\cite{sankaranarayanan2018learning} & & &\textbf{80.1} &29.1 &77.5 &2.8 &0.4 &26.8 &11.1 &18.0 &78.1 &76.7 &48.2 &15.2 &70.5 &17.4 &8.7 &16.7 &36.1  \\ \cline{2-21}
&NonAdapt~\cite{chen2018road} & \multicmark  & \multicmark  &4.7 &11.6 &62.3 &10.7 &0.0 &22.8 &4.3 &15.3 &68.0 &70.8 &49.7 &6.4 &60.5 &11.8 &2.6 &4.3 &25.4 \\ 
&ROAD~\cite{chen2018road} & &  &77.7 &30.0 &77.5 &9.6 &0.3 &25.8 &10.3 &15.6 &77.6 &79.8 &44.5 &16.6 &67.8 &14.5 &7.0 &23.8 &36.2 \\ \cline{2-21}

&NonAdapt~\cite{eccv_unsupervised} & \multicmark &\multicmark  &17.2 &19.7 &47.3 &1.1 &0.0 &19.1 &3.0 &9.1 &71.8 &78.3 &37.6 &4.7 &42.2 &9.0 &0.1 &0.9 &22.6 \\
&CBST~\cite{eccv_unsupervised} & & &69.6 &28.7 &69.5 &\textbf{12.1} &0.1 &25.4 &\textbf{11.9} &13.6 &\textbf{82.0} &\textbf{81.9} &49.1 &14.5 &66.0 &6.6 &3.7 &32.4 &35.4 \\ \cline{2-21}

&NonAdapt~\cite{wu2018dcan} &\multicmark &\multicmark  &10.8 &11.4 &66.6 &1.6 &0.1 &16.9 &5.5 &14.1 &74.2 &76.2 &46.0 &11.5 &45.4 &15.1 &6.0 &13.4 &25.9 \\ 
&DCAN~\cite{wu2018dcan} & &  &79.9 &30.4 &70.8 &1.6 &\textbf{0.6} &22.3 &6.7 &23.0 &76.9 &73.9 &41.9 &16.7 &61.7 &11.5 &\textbf{10.3} &38.6 &35.4 \\ \cline{2-21}

&BDL~\cite{li2019bidirectional} &\singlecmark &\singlecmark  &72.0 &30.3 &74.5 &0.1 &0.3 &24.6 &10.2 &\textbf{25.2} &80.5 &80.0 &\textbf{54.7} &\textbf{23.2} &\textbf{72.7} &\textbf{24.0} &7.5 &\textbf{44.9} &\textbf{39.0} \\ \cline{2-21}

&DAM~\cite{huang2018domain} & \singlecmark &\singlecmark  &-- &-- &-- &-- &-- &-- &-- &-- &-- &-- &-- &-- &-- &-- &-- &-- &30.7 \\ \cline{2-21}

&NonAdapt~\cite{zhu2018penalizing} &\multicmark &\multicmark  &-- &-- &-- &-- &-- &-- &-- &-- &-- &-- &-- &-- &-- &-- &-- &-- &24.9 \\ 
&PTP~\cite{zhu2018penalizing} & &  &-- &-- &-- &-- &-- &-- &-- &-- &-- &-- &-- &-- &-- &-- &-- &-- &34.2 \\ \cline{2-21}

&NonAdapt & \multixmark& \multicmark &15.6 &12.3 &70.3 &6.7 &0.2 &20.4 &5.6 &15.3 &73.5 &76.2 &47.2 &10.5 &54.3 &12.1 &5.3 &10.6 &27.3 \\
&Ours & &  &78.9 &\textbf{31.4} &\textbf{79.3} &9.6 &0.2 &\textbf{27.3} &10.1 &15.6 &76.2 &78.5 &45.1 &16.4 &69.8 &13.6 &8.3 &22.7 &36.4 \\ \cline{2-21}

&NonAdapt & \multixmark& \multixmark &14.7 &11.8 &68.5 &7.3 &0.1 &19.6 &4.6 &14.4 &71.8 &73.2 &48.5 &9.1 &56.1 &11.7 &4.9 &11.7 &26.8\\ 
&Ours & &  &77.5 &30.7 &78.6 &5.6 &0.2 &26.7 &10.6 &16.1 &75.2 &76.5 &44.1 &15.8 &69.9 &14.7 &8.6 &17.6 &35.5\\ \bottomrule
\end{tabular}
}
\vspace{-2mm}
\label{tab:adaptation_synthia}
\end{table*}

\end{document}